\def\year{2022}\relax
\documentclass[letterpaper]{article} 
\usepackage{aaai22}  
\usepackage{times}  
\usepackage{helvet}  
\usepackage{courier}  
\usepackage[hyphens]{url}  
\usepackage{graphicx} 
\usepackage{natbib}  
\usepackage{caption} 
\DeclareCaptionStyle{ruled}{labelfont=normalfont,labelsep=colon,strut=off} 
\usepackage{algorithm}
\usepackage{algorithmic}

\usepackage{newfloat}
\usepackage{listings}
\floatstyle{ruled}
\newfloat{listing}{tb}{lst}{}
\floatname{listing}{Listing}
\usepackage{interval}
\usepackage{bm}
\usepackage{bbm}
\usepackage{mathtools}
\usepackage{footmisc}
\usepackage{subfig}
\usepackage{xspace}
\usepackage{soul}
\usepackage{booktabs}
\usepackage{amssymb}

\newcommand{\diff}[1]{{\small{#1}}} 

\newcommand*{\eg}{\emph{e.g.}\@\xspace}

\DeclareMathOperator{\E}{\mathbb{E}}

\title{Rethinking Pseudo Labels for Semi-Supervised Object Detection}
\author{Hengduo Li$^{1}$ \qquad Zuxuan Wu$^{2}$\thanks{Corresponding author.} \qquad Abhinav Shrivastava$^{1}$ \qquad Larry S. Davis$^{1}$ \\
$^{1}$~University of Maryland \qquad $^{2}$~Fudan University \\
{\tt\small \{hdli,abhinav,lsd\}@cs.umd.edu} \quad \tt\small zxwu@fudan.edu.cn
}

\begin{document}

\maketitle

\begin{abstract}
Recent advances in semi-supervised object detection (SSOD) are largely driven by consistency-based pseudo-labeling methods for image classification tasks, producing pseudo labels as supervisory signals. However, when using pseudo labels, there is a lack of consideration in localization precision and amplified class imbalance, both of which are critical for detection tasks. In this paper, we introduce certainty-aware pseudo labels tailored for object detection, which can effectively estimate the classification and localization quality of derived pseudo labels. This is achieved by converting conventional localization as a classification task followed by refinement. Conditioned on classification and localization quality scores, we dynamically adjust the thresholds used to generate pseudo labels and reweight loss functions for each category to alleviate the class imbalance problem. Extensive experiments demonstrate that our method improves state-of-the-art SSOD performance by 1-2\% AP on COCO and PASCAL VOC while being orthogonal and complementary to most existing methods. In the limited-annotation regime, our approach improves supervised baselines by up to 10\% AP using only 1--10\% labeled data from COCO. 

\end{abstract}

\section{Introduction}
\label{sec:intro}

The astounding performance of deep neural networks on various computer vision tasks can be largely attributed to the availability of large-scale datasets that are manually labeled. However, collecting human annotations is labor-intensive and time-consuming, particularly for visual understanding tasks, like object detection~\cite{coco,openimages} and semantic segmentation~\cite{voc,cityscapes}. To remedy this, there is an ever-growing interest in semi-supervised learning (SSL), which learns feature representations with limited supervision by exploring the massive amount of unlabeled images that are readily available. While extensive studies have been conducted on SSL for image classification tasks~\cite{berthelot2019mixmatch,uda,sohn2020fixmatch,berthelot2019mixmatch,xie2020self,laine2016temporal,miyato2018virtual,bachman2014learning}, relatively limited effort has been made to address object detection, for which annotations are more expensive to obtain. 

Most recent semi-supervised object detection (SSOD) approaches~\cite{stac,unbiasedteacher,csd,instantteaching} are direct extensions of SSL methods designed for image classification using a teacher-student training paradigm~\cite{meanteacher,sohn2020fixmatch,berthelot2019mixmatch}. In particular, the teacher model is first trained in a supervised manner with a limited number of labeled samples. Then, given an unlabeled image, the teacher model produces \emph{pseudo} bounding boxes together with their corresponding class predictions, which are further used as ground-truth labels for the student model. To ensure effective distillation, the teacher and the student models typically operate on two augmented views of the same image~\cite{sohn2020fixmatch,stac,unbiasedteacher,instantteaching}. 

The use of a teacher-student model at its core aims to produce reliable pseudo labels in lieu of human annotations. While effective, we argue that pseudo labels, in the form of bounding boxes associated with class predictions, are sub-optimal for SSOD. The reasons are twofold: (1) In image classification,  prediction scores naturally represent the likelihood of an object appearing in an image, and thus setting a threshold to select highly confident predictions is reasonable. However, as detection requires localizing and classifying objects using two separate branches through regression and classification, the resulting classification scores of pseudo boxes are \emph{unaware} of the localization quality. Therefore, while widely adopted, filtering out boxes based on class predictions on top of non-maximum suppression is not appropriate; (2) Pseudo labels produced by the teacher model amplifies class imbalance which results from the long-tailed nature in detection tasks.  For example, there are only 9 toasters but 12,343 persons in 5\% of the COCO~\cite{coco} training set even though they are both \emph{common}\footnote{COCO: Common Objects in Context.} classes! As a result, lower-confidence predictions from underrepresented classes are oftentimes filtered out with a threshold that works well for top-performing classes.

\begin{figure*}[t] \centering
  \resizebox{0.95\linewidth}{!}{\includegraphics[width=\linewidth]{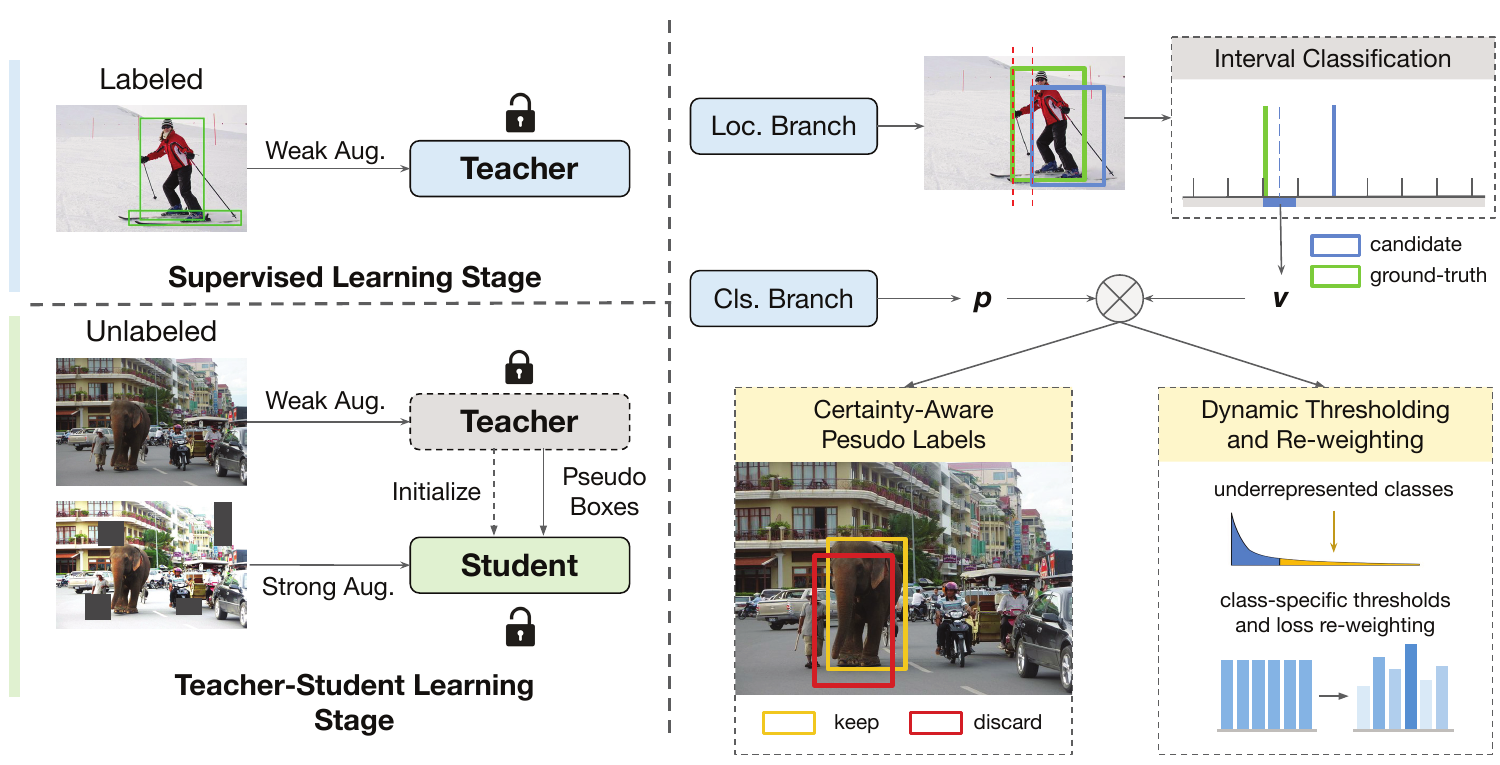}}
  \caption{\textbf{A conceptual overview of our approach.} \textbf{Left}: We first train the teacher model on labeled images to generate pseudo labels (boxes) on unlabeled images. The student model is then trained with pseudo labels. \textbf{Right}: We propose to generate certainty-aware pseudo labels conditioned on both classification and localization confidence scores, for improved localization, by formulating localization as a classification problem. The scores are then used to derive dynamic thresholds and re-weight losses in a class-wise manner to mitigate class imbalance. }
  \label{fig:approach}
\end{figure*}

To mitigate these issues, we propose certainty-aware pseudo labels together with dynamic thresholding and reweighting mechanisms tailored for SSOD. In particular, the certainty-aware pseudo labels are designed to reflect localization quality and classification confidence at the same time. Conditioned on these certainty measurements, we dynamically adjust the thresholds used to produce pseudo labels and reweight loss functions on a per-category basis to combat class imbalance. While conceptually appealing, it is challenging to have an in-vitro metric in mainstream detection frameworks that reflects localization quality to complement classification accuracy due to the design that performs localization with regression.

Motivated by a few recent studies that replace regression with classification for better localization~\cite{qiu2020offset,wang2020side}, we formulate localization as a classification problem to obtain an estimation of localization quality. More specifically, for each side of a candidate box,  we introduce a line segment that is perpendicular to it.  The line is split into consecutive intervals, each of which is associated with a prediction score through classification, indicating the probability of the side intersects with the interval. We then average the maximal classification scores from all four sides of a candidate box as its localization quality metric. To ensure accurate localization, we further refine locations within intervals. The pseudo labels are now certainty-aware, measuring both localization precision and classification confidence, and can be readily used to generate better labels. In particular, for each category, conditioned on the localization and classification confidence, we dynamically determine a threshold to generate pseudo labels and reweight loss functions such that underrepresented classes are emphasized during training to mitigate class imbalance.

We conduct extensive experiments on COCO~\cite{coco} and PASCAL VOC~\cite{voc} under common semi-supervised settings, and demonstrate that our method improves SOTA performance by 1-2\% on COCO and PASCAL VOC respectively using various training recipes while being orthogonal and complementary to most recent methods (which will be shown empirically), and improves the supervised baseline by up to 10\% AP when using only 1/2/5/10 \% annotations of COCO. We further show that our method is complementary to existing approaches resorting to orthogonal techniques like co-teaching~\cite{coteaching} and model ensemble. Extensive ablation experiments are conducted to validate the effectiveness of different components of our method, and demonstrate that our approach is relatively robust to hyper-parameter selections.

\section{Approach}
\label{sec:approach}

Our goal is to address semi-supervised object detection where a set of labeled images with box-level annotations and a set of unlabeled images are used for training. Built upon  consistency-based pseudo labeling, our method produces certainty-aware pseudo labels for both classification and localization. This is achieved by formulating box localization as a classification problem and injecting localization confidence to guide pseudo label generation. Conditioned on classification and localization certainty, we dynamically adjust the thresholds to generate pseudo labels and re-weight the loss function  for different classes. An overview of our method is shown in Fig.~\ref{fig:approach}.

\subsection{Preliminary} \label{sec:approach_prelim}
Our approach is built upon consistency-based pseudo labeling, which has proven effective for both semi-supervised image classification and object detection. Below, we briefly introduce the teacher-student training paradigm which serves as the basis for current consistency-based approaches. Overall, a teacher model is firstly trained on labeled images, and then it is used to produce pseudo labels (boxes) on unlabeled images to supervise the training of a student model.

Formally, given a set of labeled images $\mathcal{S}$ and  a set of unlabeled images $\mathcal{U}$, an object detector is trained on $\mathcal{S}$ in a standard supervised manner:
\begin{align} 
{\bf \mathcal{L}}_s(\mathcal{I}, {\bf p}, {\bf t}, {\bf p}_{*}, {\bf t}_{*}) & = \E_{\mathcal{I} \sim \mathcal{S}} \E_{i \sim \mathcal{B}} \mathcal{L}_s(\mathcal{I}, p^i, t^i) \nonumber\\
&=  \E_{\mathcal{I} \sim \mathcal{S}} \E_{i \sim \mathcal{B}} [\ell_{cls}(p^i, p^i_*) \ + \ell_\text{loc}(t^i, t^i_*) ]
\label{eq:loss_sup}
\end{align} 

where $\mathcal{I}$ is an input image with a set of candidate boxes $\mathcal{B}$, and $p^i$, $t^i$ denote the prediction of class probability and bounding box coordinates for the $i$-th candidate box. Each candidate box is associated with a one-hot label $p^i_*$ and a ground-truth box location $t^i_*$ as supervisory signals, and the losses for classification and localization are often instantiated as a weighted sum of a standard cross-entropy loss and a smooth L$_{1}$.

The teacher model trained on $\mathcal{S}$ then generates pseudo boxes on all unlabeled images in $\mathcal{U}$ through standard inference. These pseudo boxes are further filtered by a predefined threshold $\tau$ conditioned on the prediction confidence $p^i$; the remaining boxes are used to train a student model whose weights are initialized from the teacher model: 
\begin{align} 
{\bf \mathcal{L}} = \mathcal{L}_s(\mathcal{I}_s, {\bf p}_*, {\bf t}_*) + \lambda_u \mathcal{L}_u(\mathcal{I}_u, {\bf p}_{u*}, {\bf t}_{u*})
\label{eq:loss_unsup}
\end{align} 
where ${\bf p}_{u*}$ and ${\bf t}_{u*}$ denote pseudo class labels and box coordinates derived from the teacher model. The loss is a weighted sum of supervised loss $\mathcal{L}_s$ on labeled images and unsupervised loss $\mathcal{L}_u$ on unlabeled samples controlled by $\lambda$. Following~\cite{stac,unbiasedteacher,instantteaching,ismt}, given an unlabeled image, when generating pseudo labels, we only use horizontal flipping as a weak augmentation; when training the student model, we use strong augmentations including color jitter, Gaussian blur and Cutout~\citeyearpar{cutout} for the same image.

\begin{table*}[t]  \centering
\subfloat[\emph{COCO-full}]{
    \resizebox{!}{2.2cm}{\begin{tabular}{@{}lc@{}}
    \toprule
    Method & AP\\
    \midrule
    Supervised & 37.9 \\
    Supervised$^*$ & 40.2 \\
    STAC$^*$~\citeyearpar{stac} & 39.2 \\
    ISMT~\citeyearpar{ismt} & 39.6 \\
    Instant-Teaching~\citeyearpar{instantteaching} & 39.6 \\
    Multi Phase$^*$~\citeyearpar{multiphaselearning} & 40.1 \\
    Unbiased Teacher$^*$~\citeyearpar{unbiasedteacher} & 41.3 \\
    Humble teacher$^*$~\citeyearpar{tang2021humble} & 42.3 \\
    \midrule
    \textbf{Ours} & \textbf{41.0} \\
    \textbf{Ours}$^*$  & \textbf{43.3} \\
    \bottomrule
    \end{tabular}}}
\hspace{4mm}
\subfloat[\emph{Pascal VOC}.]{
    \resizebox{!}{2.2cm}{\begin{tabular}{@{}lccc@{}}
    \toprule
    Method &  AP$_{50}$ & AP$_{75}$ & AP \\
    \midrule
    Supervsied & 76.3 & 47.5 & 45.3 \\
    CSD~\citeyearpar{csd} & 74.7 & - & - \\
    STAC~\citeyearpar{stac} & 77.4 & - & 44.6 \\
    ISMT~\citeyearpar{ismt} & 77.2 & - & 46.2 \\
    Instant-Teaching~\citeyearpar{instantteaching} & 78.3 & 52.0 & 48.7\\
    Multi Phase~\citeyearpar{multiphaselearning} & 77.4 & - & - \\
    Unbiased Teacher$^\mathsection$~\citeyearpar{unbiasedteacher} & 77.4 & - & 48.7 \\
    Humble teacher$^\mathsection$~\citeyearpar{tang2021humble} & 80.9 & - & 53.0 \\
    \midrule
    \textbf{Ours} & \textbf{76.9} & \textbf{57.9} & \textbf{52.4} \\
    \textbf{Ours}$^\mathsection$  & \textbf{79.0} & \textbf{59.4} & \textbf{54.6} \\
    \bottomrule
    \end{tabular}}}
\hspace{4mm}
\subfloat[\emph{Pascal VOC + COCO-20}.]{
    \resizebox{!}{2.2cm}{\begin{tabular}{@{}lccc@{}}
    \toprule
    Method &  AP$_{50}$ & AP$_{75}$ & AP \\
    \midrule
    Supervsied & 76.3 & 47.5 & 45.3 \\
    CSD~\citeyearpar{csd} & 75.1 & - & - \\
    STAC~\citeyearpar{stac} & 79.1 & - & 46.0 \\
    ISMT~\citeyearpar{ismt} & 77.7 & - & 49.6 \\
    Instant-Teaching~\citeyearpar{instantteaching} & 79.0 & 54.1 & 49.7 \\
    Multi Phase~\citeyearpar{multiphaselearning} & - & - & - \\
    Unbiased Teacher$^\mathsection$~\citeyearpar{unbiasedteacher} &  78.8 & - & 50.3 \\
    Humble teacher$^\mathsection$~\citeyearpar{tang2021humble} & 81.3 & - & 54.4 \\
    \midrule
    \textbf{Ours} & \textbf{77.6} & \textbf{59.1} & \textbf{54.0} \\
    \textbf{Ours}$^\mathsection$  & \textbf{79.6} & \textbf{61.2} & \textbf{56.1} \\
    \bottomrule
    \end{tabular}}}
\caption{\textbf{Comparison with state-of-the-art approaches on \emph{COCO-full}, \emph{PASCAL VOC}, and \emph{PASCAL VOC + COCO-20} settings.} $^*$ denotes the use of longer training schedule (3$\times$). $^\mathsection$ denotes multi-scale training.}
\label{tab:main_full}
\end{table*}

\subsection{Certainty-aware Pseudo Labels} \label{sec:approach_loc}

Recall that existing approaches typically form bounding boxes through coordinate regression, and then predict the objects within boxes through classification.  To generate pseudo boxes used as ground-truth by the student model, it is a common practice to apply a threshold $\tau$ to filter out boxes with low classification scores. While straightforward, such a localization-agnostic strategy fails to model how well boxes are localized. To address this issue, we formulate localization as classification, producing certainty-aware boxes, such that the quality of both localization and classification are explicitly considered to guide the generation of pseudo labels.

Formally, given an \emph{unlocalized} candidate box $(x_1, y_1, x_2, y_2)$ with its top left corner at $(x_1, y_1)$ and its bottom right corner at $(x_2, y_2)$, its corresponding ground-truth locations are denoted as $(x_{g1}, y_{g1}, x_{g2}, y_{g2})$. Each side of the candidate is independently localized to the corresponding side of ground-truth through \emph{classification}. Taking the left side of the candidate box as an example, we first obtain a line segment $l$ which is perpendicular to the side, then split $l$ evenly into $K$ consecutive intervals and predict which interval the \emph{unlocalized} side belongs to according to the ground-truth position $x_{g1}$ through a $K$-way classification. In particular, if the left side of the ground-truth box is perpendicular to the $k$-th interval, we mark that the side belongs to the $k$-th interval for training (see Figure~\ref{fig:approach} for an illustration). Then the loss function for localization given an image can be written as:
\begin{align} 
{\bf \ell}_\text{seg}({\bm T}, {\bm Y}) &= \E_{i \sim \mathcal{B}} \sum_{s=1}^{s=4}\sum_{k=1}^{k=K} - y_{s,k}^i \ \texttt{log}(\texttt{sigmoid}(t_{s,k}^i))
\label{eq:loss_int}
\end{align} 
where the superscript $i$ denotes the $i$-th candidate box sampled from the box set, and $t_{k,s}^i$ is the unnormalized prediction score from the $k$-th interval in the $s$-th side, and the label $y_{k,s}^i = 1$ if the side belongs to the $k$-th interval otherwise $y_{k,s}^i = 0$. To measure the localization quality for the $i$-th box, we first obtain the maximal class score along each side and then compute the mean of these scores:
\begin{align}
    v^i = \frac{1}{4} \sum_{s=1}^{s=4} \texttt{max}_{1 \leq k \leq K}(t_{s,k}^i).
\end{align}
The localization quality score $v^i$, indicating how well boxes are localized, together with the classification confidence $p^i$ are two complementary metrics measuring the certainty of localization and classification, respectively. They are further used for post-processing like non-maximum suppression and pseudo label generation, which will be described below.

Thus far we have formulated box localization in a classification manner to obtain quality measurement, yet the localization performance could be largely hindered by discretizing the problem of deriving continuous bounding box coordinates. In particular, the membership of an interval is a rough estimation of location particularly when the interval size is large. 
To obtain the precise location of a side within the interval, we further perform regression from the center line $x_k$ of the $k$-th interval to the ground-truth line $x_g$ for finer localization. We use a smooth $L_1$ loss for fine regression, and the overall localization loss becomes:
\begin{align} 
{\bf \ell}_\text{loc} &= {\bf \ell}_\text{seg}({\bm T}, {\bm Y}) + {\bf \ell}_\text{reg}({\bm Y}, {\bm X}_g) \nonumber\\
&= \E_{i\sim \mathcal{B}} \sum_k^K y_k \ [\ - \texttt{log}(\texttt{sigmoid}(t_{k,s}^i)) \nonumber \\
&\quad + \texttt{SmoothL1}(x_k, x_g)\ ] 
\label{eq:loss_all}
\end{align} 
Finally, we replace the localization loss in Eqn.~\ref{eq:loss_sup} used by both the teacher model and the student model with Eqn.~\ref{eq:loss_all}. Consequently, the trained teacher model produces pseudo labels that are aware of both localization and classification quality. 

\subsection{Dynamic Thresholding and Re-weighting} \label{sec:approach_dynamic}
As discussed above, class imbalance exists in object detection especially when annotations are scarce. The imbalance is further enlarged in semi-supervised settings since the teacher model produces relatively lower confidence scores for underrepresented classes~\cite{dave2021evaluating}, which hardly survive the often large threshold $\tau$. On the other hand, simply lowering $\tau$ introduces more noisy pseudo labels in common classes. With this in mind, we propose to dynamically adjust the threshold and re-weight losses in a class-wise manner conditioned on classification and localization confidence scores for each category.

For each category $m$, the classification and localization confidence score $p_m^j$ and $v_m^j$ for each foreground candidate box (indexed by $j$) are accumulated online to produce an unnormalized frequency score $c = \sum_j p_m^j v_m^j$, which not only approximates the detector's current overall confidence level for the category but also counts the number of foreground instances. The class-specific threshold $\tau_m$ and re-weighting coefficient $\alpha_m$ are then derived as follows:
\begin{align} 
\tau_m = \left(\frac{\sum_j p_m^j v_m^j}{\left.\E_{m\sim M} \sum_j \mathbbm{1}\right.}\right)^{\gamma_1} \tau, \quad \alpha_m = \left(\frac{\left.\E_{m \sim M} \sum_j \mathbbm{1}\right.}{\sum_j p_m^j v_m^j}\right)^{\gamma_2}
\end{align} 
where $\E_{m \sim M} \sum_j \mathbbm{1}$ denotes the average number of foreground instances from all categories and $\tau$ is the original manually chosen threshold. The class-specific $\tau_m$ is then applied to filter pseudo labels, and $\alpha_m$ is multiplied to losses (Eqn.~\ref{eq:loss_unsup}) of all foreground instances in each category. Two factors $\gamma_1$ and $\gamma_2$ control the degree of focus on underrepresented classes; when set to $0$, dynamic thresholding and re-weighting are disabled.

By keeping more pseudo labels for underrepresented classes, as well as promoting their importance during training through re-weighting the losses, the bias towards head classes is mitigated. It is worth pointing out that $\tau_m$ needs to be bounded as it is applied on predicted probabilities, and we find clipping it into $[0.4, 0.9]$ works well empirically.

\section{Experiments} \label{sec:exp}

\subsection{Experimental Setup} \label{sec:exp_setup}

\noindent\textbf{Datasets.} We evaluate our method on two standard object detection datasets, COCO~\cite{coco} and PASCAL VOC~\cite{voc}, under semi-supervised settings following~\cite{csd,stac,unbiasedteacher,instantteaching,ismt}. In particular, four settings are used: (1) \emph{COCO-full}: the COCO \texttt{train2017} set containing $\sim$118k images is used as the labeled set, and the additional $\sim$123k unlabeled images are used as unlabeled set; (2) \emph{COCO-partial}: we follow~\cite{stac} and randomly sample $1\%/2\%/5\%/10\%$ images from COCO \texttt{train2017} set as the labeled set, and use the remaining images in \texttt{train2017} as the unlabeled set; (3) \emph{PASCAL VOC}: the \texttt{VOC07 trainval} set is used as labeled set and the \texttt{VOC12 trainval} is used as unlabeled set; (4) \emph{PASCAL VOC + COCO-20}: following~\cite{stac}, images from COCO containing the 20 classes in PASCAL VOC are used as an additional unlabeled set. For evaluation, the \texttt{val2017} set of COCO and the \texttt{VOC07 test} set of PASCAL VOC are used.

\noindent\textbf{Training and Testing Configuration.} Since existing methods for SSOD use various different setups for training and testing, we evaluate our method under multiple settings for fair comparison. In all settings, the teacher model is firstly trained on the  labeled set, and the student model is trained on the combination of labeled and unlabeled images. We report mean Average Precision (mAP) at different IoU thresholds (\eg AP$_{50}$, AP$_{75}$ and AP$_{50:95}$ which is denoted as AP) to measure the effectiveness. 


\noindent\textbf{Implementation Details.} Our implementation follows existing approaches for fair comparison, and thus we use Faster-RCNN with FPN~\cite{fpn} as our detector using a ResNet-50~\cite{resnet} as its backbone network. 


For more details of the implementation such as choices of hyper parameters and training recipes, we refer readers to Appendix.

\begin{table*}[t]  \centering
    \resizebox{0.9\linewidth}{!}{\begin{tabular}{@{}lcccc@{}}
    \midrule
    Methods & $1\%$ COCO & $2\%$ COCO & $5\%$ COCO & $10\%$ COCO \\
    \midrule
    Supervised & 9.05 $\pm$ 0.16 & 12.70 $\pm$ 0.15 & 18.47 $\pm$ 0.22 & 23.86 $\pm$ 0.81  \\
    CSD~\citeyearpar{csd} & 10.20 $\pm$ 0.15 \diff{(+1.15)} & 13.60 $\pm$ 0.10 \diff{(+0.90)} & 18.90 $\pm$ 0.10 \diff{(+0.43)} & 24.50 $\pm$ 0.15 \diff{(+0.64)} \\
    STAC~\citeyearpar{stac} & 13.97 $\pm$ 0.35 \diff{(+4.92)} & 18.25 $\pm$ 0.25 \diff{(+5.55)} & 24.38 $\pm$ 0.12 \diff{(+5.91)} & 28.64 $\pm$ 0.21 \diff{(+4.78)} \\
    Unbiased Teacher~\citeyearpar{unbiasedteacher} & 17.84 $\pm$ 0.12 \diff{(+8.79)} & 21.98 $\pm$ 0.07 \diff{(+9.28)} & 26.30 $\pm$ 0.11 \diff{(+7.83)} & 29.64 $\pm$ 0.10 \diff{(+5.78)} \\
    Humble teacher$_\ddagger$~\citeyearpar{tang2021humble} & 16.96 $\pm$ 0.38 \diff{(+7.91)} & 21.72 $\pm$ 0.24 \diff{(+9.02)} & 27.70 $\pm$ 0.15 \diff{(+9.23)} & 31.61 $\pm$ 0.28 \diff{(+7.74)} \\
    Instant-Teaching$^\dagger$~\citeyearpar{instantteaching} & 16.00 $\pm$ 0.20 \diff{(+6.95)} & 20.70 $\pm$ 0.30 \diff{(+8.00)} & 25.50 $\pm$ 0.05 \diff{(+7.03)} & 29.45 $\pm$ 0.15 \diff{(+5.59)} \\
    Instant-Teaching$^\dagger_\ddagger$~\citeyearpar{instantteaching} & 18.05 $\pm$ 0.15 \diff{(+9.00)} & 22.45 $\pm$ 0.15 \diff{(+9.75)} & 26.75 $\pm$ 0.05 \diff{(+8.28)} & 30.40 $\pm$ 0.05 \diff{(+6.54)} \\
    \midrule
    \textbf{Ours} & \textbf{18.21 $\pm$ 0.31 \diff{(+9.16)}} & \textbf{22.62 $\pm$ 0.24 \diff{(+9.92)}} & \textbf{27.78 $\pm$ 0.17 \diff{(+9.31)}} & \textbf{31.67 $\pm$ 0.18 \diff{(+7.81)}} \\
    \textbf{Ours}$^\dagger$ & \textbf{19.02 $\pm$ 0.25 \diff{(+9.97)}} & \textbf{23.34 $\pm$ 0.18 \diff{(+10.64)}} & \textbf{28.40 $\pm$ 0.15 \diff{(+9.93)}} & \textbf{32.23 $\pm$ 0.14 \diff{(+8.37)}} \\
    \bottomrule
\end{tabular}}
\caption{\textbf{Results (AP) on \emph{COCO-partial}.} $\dagger$ denotes using a lower final score threshold to improve recall as in~\cite{instantteaching}. $\ddagger$ denotes using ensemble.}
\label{tab:main_partial}
\end{table*}

\begin{table*}[!t]  \centering
\begin{minipage}[b]{0.28\linewidth} \centering
\resizebox{\linewidth}{!}{
\begin{tabular}{@{}cc@{}}
    \toprule
    2\% COCO & AP \\
    \midrule
    Overall & 21.6 $\longrightarrow$ \textbf{22.5} \\
    Rarest 10 Classes & 23.9 $\longrightarrow$ \textbf{26.0} \\
    \bottomrule
\end{tabular}}
\caption{Performance improvement on the rarest 10 classes.}
\label{tab:main_imbalance}
\end{minipage}
\quad
\begin{minipage}[b]{0.3\linewidth} \centering
\resizebox{\linewidth}{!}{
\begin{tabular}{@{}cccc@{}}
    \toprule
    2\% COCO & AP$_{50}$ & AP$_{75}$ & AP \\
    \midrule
    Single Model & 37.1 & 23.7 & 22.5 \\
    Ensemble & \textbf{37.9} & \textbf{24.1} & \textbf{23.0} \\
    \bottomrule
\end{tabular}}
\caption{Performance of our method with model ensemble.}
\label{tab:main_complementary}
\end{minipage}
\quad
\begin{minipage}[b]{0.33\linewidth} \centering
\resizebox{\linewidth}{!}{
\begin{tabular}{@{}cccc@{}}
    \toprule
    VOC + COCO-20 & AP$_{50}$ & AP$_{75}$ & AP \\
    \midrule
    Original & 79.6 & 61.2 & 56.1 \\
    1:1 Sampling & \textbf{79.8} & \textbf{62.1} & \textbf{56.9} \\
    \bottomrule
\end{tabular}}
\caption{Performance with 1:1 labeled:unlabeled image sampling ratio.}
\label{tab:main_complementary_2}
\end{minipage}
\end{table*}

\subsection{Main Results}

We first report the results on four settings, and compare with supervised baselines as well as various state-of-the-art approaches for semi-supervised object detection, such as CSD~\citeyearpar{csd}, STAC~\citeyearpar{stac}, ISMT~\citeyearpar{ismt}, Instant-Teaching~\citeyearpar{instantteaching}, Multi-Phase Learning~\citeyearpar{multiphaselearning}, Unbiased Teacher~\citeyearpar{unbiasedteacher} and Humble teacher~\citeyearpar{tang2021humble}. For approaches using ensemble techniques like~\cite{instantteaching,multiphaselearning}, we report their single-model results for fair comparison. For Unbiased Teacher~\cite{unbiasedteacher} which uses larger batch size and longer training schedules, we retrain it under our training schedules with their official implementation. Results are summarized in Table~\ref{tab:main_full} and Table~\ref{tab:main_partial}.

\noindent\textbf{\emph{COCO-full} and \emph{PASCAL VOC}.} As shown in Table~\ref{tab:main_full}(a-c), our approach outperforms state-of-the-art methods by at least $1-2\%$ AP on COCO and PASCAL VOC. For example, when trained under 3$\times$ schedule, our method obtains $43.3\%$ AP and outperforms Unbiased Teacher~\citeyearpar{unbiasedteacher} and Humble teacher~\citeyearpar{tang2021humble} by $2.0\%$ and $1.0\%$ respectively. In the short schedule (1$\times$) setting, our approach obtains $41.0\%$ AP, which outperforms methods using long schedules like CSD~\citeyearpar{csd} and STAC~\citeyearpar{stac}. On PASCAL VOC, we obtain $52.4\%$ AP and $54.6\%$ AP with single-scale training and multi-scale training respectively. Notably, large improvements are obtained by our method when precise localization is needed (\eg AP$_{75}$), indicating that our approach improves the localization quality for SSOD. 

\noindent\textbf{\emph{COCO-partial}.} We then evaluate our method under the limited-annotation regime on \emph{COCO-partial}. As demonstrated in Table~\ref{tab:main_partial},  our method improves supervised baselines by up to $10\%$. When $10\%$ annotations are available, our method achieves $32.23\%$ AP and is $\sim2\%$ higher than Instant-Teaching~\cite{instantteaching} even though model ensemble is used in their method. With only $1\%/2\%/5\%$ annotations available, our method is able to achieve state-of-the-art $19.02\%$, $23.34\%$ and $28.40\%$ APs respectively.

\noindent\textbf{Improvements for underrepresented classes.} To validate the effectiveness of our method on improving the detection performance for underrepresented classes, we also show results (Table~\ref{tab:main_imbalance}) on the rarest 10 classes in terms of number of annotations in the training set. We can see after adding dynamic pseudo label thresholding and loss re-weighting methods described in Sec.~\ref{sec:approach_dynamic}, the overall performance is improve by 0.9\% AP (from 21.6\% to 22.5\%) and the performance on rare classes is improved by 2.1\% AP (from 23.9\% to 26.0\%). This confirms that our method indeed promotes the performance for underrepresented classes. 

\noindent\textbf{Compatibility to other methods.} It is worth pointing out that our method is orthogonal to many useful techniques explored in existing approaches mentioned above. For example, when using a simplified model ensemble method from~\cite{instantteaching}, a further performance improvement is observed as in Table~\ref{tab:main_complementary}. In particular, we train two teacher models separately and use the ensemble of them to generate pseudo labels, which are then used to train two student models. Finally, the ensemble of two student models is evaluated. As can be seen, AP$_{50}$ and overall AP are improved by $0.8\%$ and $0.5\%$ respectively. 

We also show in Table~\ref{tab:main_complementary_2} that sampling labeled and unlabeled images with 1:1 ratio during training as in~\cite{unbiasedteacher,tang2021humble} further improves performance of our method.
Other methods like Mean Teacher~\citeyearpar{meanteacher}, Co-teaching~\citeyearpar{coteaching}, input ensemble and soft labels have also been utilized in~\cite{unbiasedteacher,instantteaching,ismt,tang2021humble} but not in our method, for which we believe our work could be complementary to many current state-of-the-art methods for semi-supervised object detection --- and thus the performance could be further improved when combining these methods with ours.

\begin{figure*}[!t] \centering
\begin{minipage}[b]{0.4\linewidth} \centering
  \resizebox{\linewidth}{!}{\includegraphics[width=\linewidth]{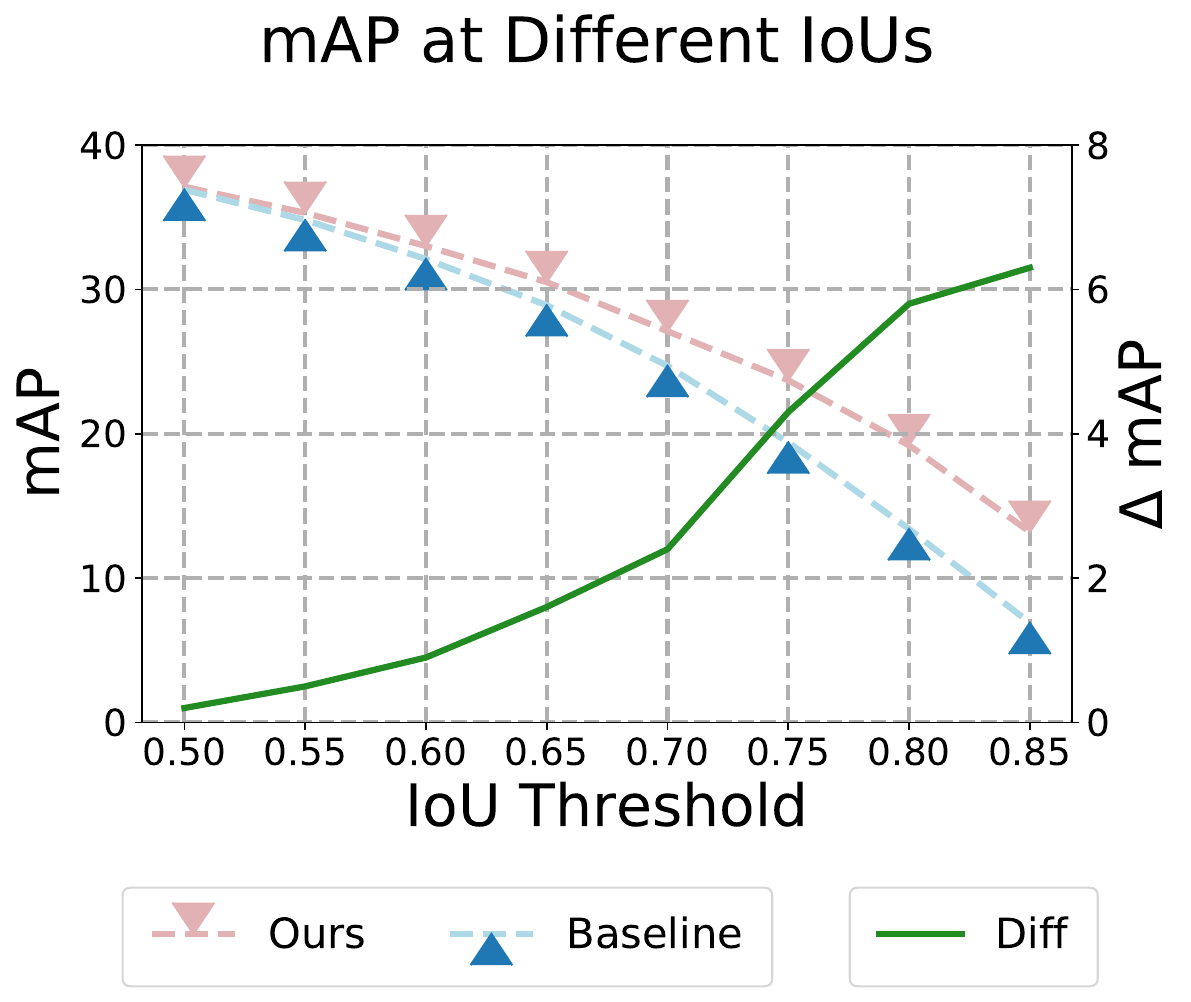}}
  \caption{\textbf{Performance at different IoU criteria} under 2\% COCO setting. }
  \label{fig:map_iou}
\end{minipage}
\hspace{1.6cm}
\begin{minipage}[b]{0.4\linewidth} \centering
  \resizebox{\linewidth}{!}{\includegraphics[width=\linewidth]{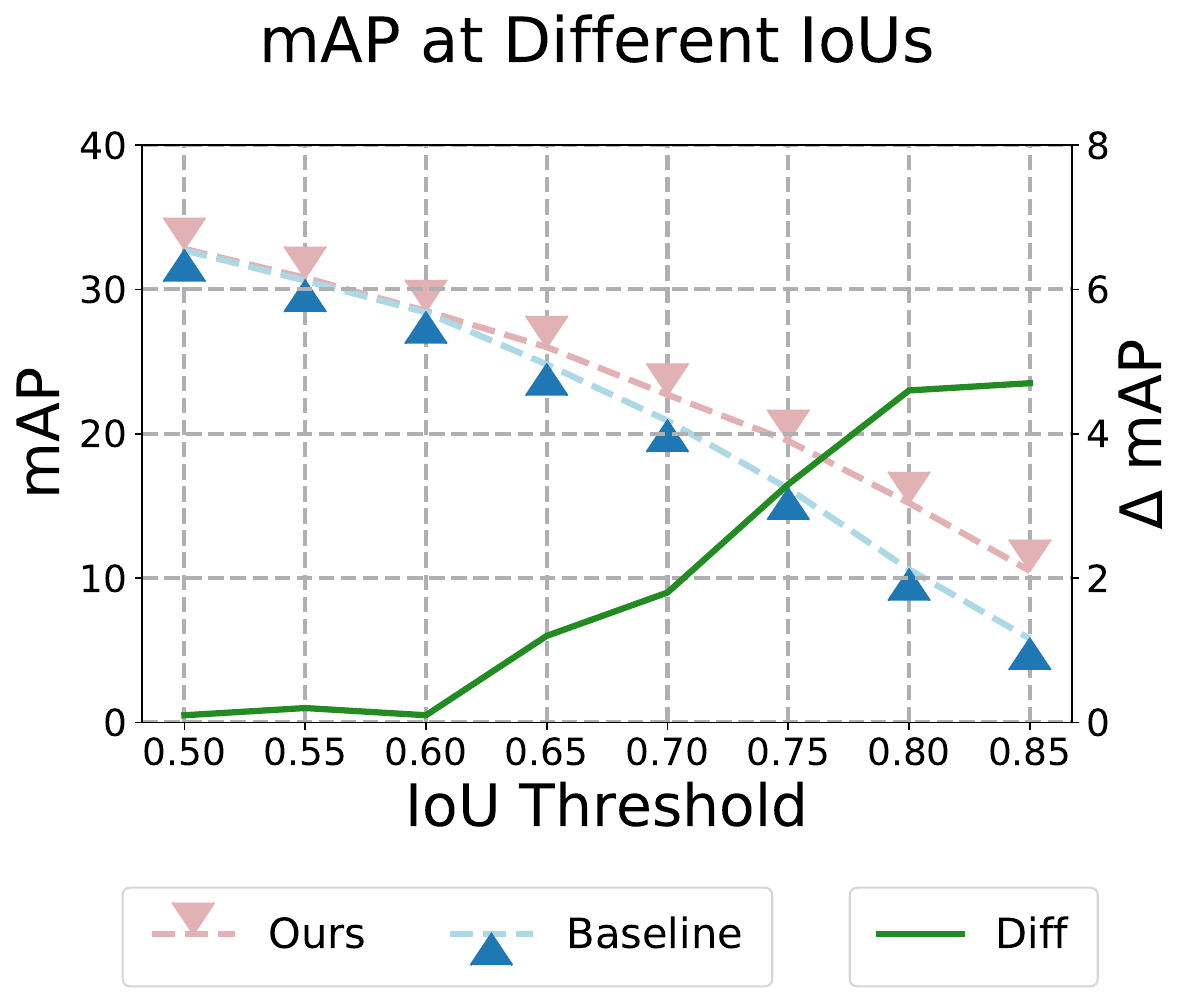}}
  \caption{\textbf{Evaluating the localization precision of pseudo boxes} from teacher model. } 
  \label{fig:loc_pl}
\end{minipage}
\end{figure*}

\noindent\textbf{Improvement on localization performance.} Having demonstrated the overall efficacy of our approach, we now evaluate the localization performance. We first compare our method against the baseline without the proposed components at different IoU criteria. As shown in Figure~\ref{fig:map_iou}, our method improves baseline by a larger margin when higher localization precision is required: the performances are similar at $0.5$ IoU threshold, whereas our approach obtains more than $6\%$ higher mAP at $0.85$ IoU threshold. We further evaluate the performance of teacher models on the withheld unlabeled images and see whether pseudo labels produced by our method are better localized. Similar trends in Figure~\ref{fig:loc_pl} confirm that pseudo labels produced by our method are localized more precisely, and thus improve the detection performance for semi-supervised object detection.

\begin{figure*}[t] \centering
  \resizebox{0.98\linewidth}{!}{\includegraphics[width=\linewidth]{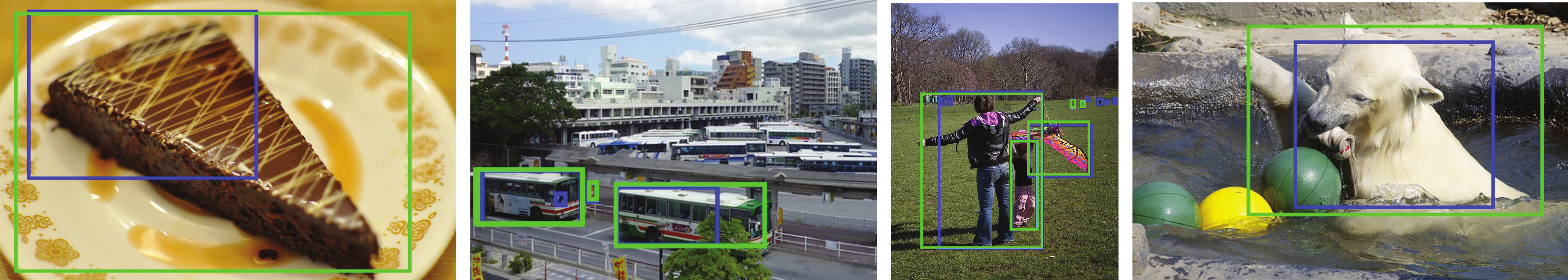}}
  \caption{\textbf{Visualization of localization quality}. Our method (Green) localizes objects more precisely than the baseline (Blue).}
  \label{fig:vis}
\end{figure*}

\noindent\textbf{Qualitative results.} In addition to the quantitative analysis presented above, we provide some qualitative results in Figure~\ref{fig:vis}. As can be observed, our method produces more precise localization results than the baseline without proposed components in Sec.~\ref{sec:approach_loc}. In particular, our method is better at localizing boundaries of irregular-posed objects like the bear and person in Figure~\ref{fig:vis}.

\subsection{Ablation Study}

\begin{table}[b]  \centering
\resizebox{0.8\linewidth}{!}{
    \begin{tabular}{@{}cccccccc@{}}
    \toprule
    CA & RE & DT &&& AP$_{50}$ & AP$_{75}$ & AP \\
    \midrule
    & & &&& 36.9 & 19.4 & 19.9 \\
    \checkmark & & &&& 35.3 & 22.9 & 21.6 \\
    & \checkmark  & &&& \textbf{38.6} & 19.9 & 20.7 \\
    & & \checkmark &&& 38.4 & 19.6 & 20.4 \\
    \checkmark & \checkmark & \checkmark &&& 37.1 & \textbf{23.7} & \textbf{22.5} \\
    \bottomrule
    \end{tabular}}
\caption{\textbf{Effectiveness of proposed components} including certainty-aware pseudo labels (\textbf{CA}), loss re-weighting (\textbf{RE}) and dynamic thresholding (\textbf{DT}).}
\label{tab:ablation_components}
\end{table}

\noindent\textbf{Effectiveness of different components.} We validate the effectiveness of proposed components and summarize the results in Table~\ref{tab:ablation_components}. We can see by adding the certainty-aware pseudo labels, class-specific loss re-weighting and dynamic thresholding, the performance is improved by $1.7\%$, $0.8\%$ and $0.5\%$ respectively. When all the components are added, our approach improves the baseline by $2.6\%$ AP and $4.3\%$ AP$_{75}$, confirming the proposed components are effective and especially useful for improving localization quality.

\noindent\textbf{Data augmentations.} We also study the usefulness of different data augmentation techniques. Table~\ref{tab:ablation_augmentation} summarizes the results. When no data augmentation is applied for training the student model, the performance degrades from $22.5\%$ to $20.3\%$ AP, indicating that data augmentation is critical. Adding color jittering and Gaussian blurring improves the performance by $1.2\%$, and applying Cutout further boosts AP by $1\%$. 

\begin{table}[!b]  \centering
\resizebox{0.85\linewidth}{!}{
    \begin{tabular}{@{}ccccccc@{}}
    \toprule
    Color & Blur & Cutout && AP$_{50}$ & AP$_{75}$ & AP \\
    \midrule
    & & && 33.9 & 21.3 & 20.3 \\
    \checkmark & & && 34.9 & 22.4 & 21.1 \\
    \checkmark & \checkmark & && 35.5 & 22.7 & 21.5 \\
    \checkmark & \checkmark & \checkmark && \textbf{37.1} & \textbf{23.7} & \textbf{22.5} \\
    \bottomrule
    \end{tabular}}
\caption{\textbf{Effectiveness of different data augmentations} applied when training the student model, including color jitter (\textbf{Color}), Gaussian blur (\textbf{Blur}) and \textbf{Cutout}.}
\label{tab:ablation_augmentation}
\end{table}

\begin{table}[!b]  \centering
\begin{minipage}[b]{0.42\linewidth} \centering
\resizebox{\linewidth}{!}{
    \begin{tabular}{@{}ccccc@{}}
    \toprule
    $K$ && AP$_{50}$ & AP$_{75}$ & AP \\
    \midrule
    4 && 37.2 & 18.9 & 19.7 \\
    8 && 38.1 & 21.1 & 21.0 \\
    20 && 36.3 & 23.4 & 22.3 \\
    30 && 37.1 & 23.7 & 22.5 \\
    40 && 37.9 & 23.4 & 22.4 \\
    \bottomrule
    \end{tabular}}
\caption{Hyper parameter sensitivity on number of intervals $K$.}
\label{tab:ablation_hyperparam1}
\end{minipage}
\hspace{2mm}
\begin{minipage}[b]{0.53\linewidth} \centering
\resizebox{\linewidth}{!}{
    \begin{tabular}{@{}cccccc@{}}
    \toprule
    $\gamma_1$ & $\gamma_2$ && AP$_{50}$ & AP$_{75}$ & AP \\
    \midrule
    0.05 & 0.4 && 36.6 & 23.2 & 22.2 \\
    0.05 & 0.6 && 37.1 & 23.7 & 22.5 \\
    0.05 & 0.8 && 37.0 & 23.7 & 22.5 \\
    0.03 & 0.6 && 36.8 & 23.6 & 22.3 \\
    0.07 & 0.6 && 36.7 & 23.3 & 22.3 \\
    \bottomrule
    \end{tabular}}
\caption{Hyper-parameter sensitivity on variance controlling factors $\gamma_1$ and $\gamma_2$.}
\label{tab:ablation_hyperparam2}
\end{minipage}
\end{table}

\noindent\textbf{Hyper-parameter Sensitivity.} We experiment with different hyper-parameters and summarize the results in Table~\ref{tab:ablation_hyperparam1} and~\ref{tab:ablation_hyperparam2}. For localization, our method is robust to hyper-parameter selection as long as $K$ is large enough to produce fine-grained localization intervals. However, when $K$ is set to be a small number like $4$, the intervals are too coarse and the localization branch degenerates to a similar form of pure regression method, resulting in a degraded performance of $19.7\%$ AP. For dynamic thresholding and loss re-weighting, larger $\gamma_1$ and $\gamma_2$ leads to more emphasis on infrequent classes during training, and we find using $\gamma_1 = 0.05$ and $\gamma_2 = 0.6$ gives the best result, as shown in Table~\ref{tab:ablation_hyperparam2}. When set as $0$, the corresponding method is disabled.

\section{Related Work}

\noindent\textbf{Object Detection.} As a fundamental computer vision task, object detection has been extensively studied for  decades~\cite{viola,dpn,faster,yolo,detr}. Modern object detectors have evolved from anchor-based detectors like Faster RCNN~\citeyearpar{faster}, YOLO~\citeyearpar{yolo} and SSD~\citeyearpar{ssd}, to anchor-free~\cite{fcos,centernetzhou} and transformer-based detectors~\cite{detr} in pursuit of simpler formulation and stronger performance. Various directions have also been actively explored on improving localization precision~\cite{iounet,wang2020side,qiu2020offset}, inference efficiency~\cite{najibi2019autofocus,uzkent2020efficient}, training paradigms~\cite{atss,freeanchor,noisyanchor}, to name a few. While powering a wide range of applications, standard object detectors require box annotations for all objects-of-interest in images during training, which are time-consuming and labour-intensive to obtain. 

\noindent\textbf{Semi-Supervised Learning.} Semi-supervised learning (SSL) for visual understanding leverages unlabeled images for improved performance in various tasks~\cite{berthelot2019mixmatch,uda,sohn2020fixmatch,miyato2018virtual,bachman2014learning,xie2020self,laine2016temporal}. Recent advances on SSL mostly resort to consistency-based methods with data augmentation and have significantly improved performance for image classification. Specifically, the model is incentivized to produce consistent predictions across different views of an input image generated with semantics-preserving data augmentations. Typical approaches like MixMatch~\citeyearpar{berthelot2019mixmatch} and UDA~\citeyearpar{uda} enforce a consistent prediction of class distributions across multiple views, while FixMatch~\citeyearpar{sohn2020fixmatch} encourages correct predictions on strongly augmented unlabeled images given one-hot pseudo labels generated on weakly augmented ones. Data augmentations used in existing methods span conventional techniques~\cite{cutout,zhang2017mixup}, learned augmentations~\cite{cubuk2018autoaugment,cubuk2020randaugment} and adversarially generated ones~\cite{miyato2018virtual}. A line of work following Mean Teacher~\citeyearpar{meanteacher} also explore updating teacher model with an Exponential Moving Average (EMA) of student model during SSL training to provide better pseudo labels~\cite{sohn2020fixmatch,cai2021exponential}. Our work follows the consistency-based paradigm with a focus on object detection, which is relatively under-explored compared to image classification yet it requires fine grained annotations.

\noindent\textbf{Semi-Supervised Object Detection.} 
The expensive labeling cost of object detection has also drawn a growing attention on developing effective SSL methods. CSD~\citeyearpar{csd} enforces consistent predictions on original and horizontally flipped images, whereas STAC~\citeyearpar{stac} encourages consistency between weakly and strongly augmented views of images as in FixMatch~\citeyearpar{sohn2020fixmatch}. On top of them, methods like Unbiased Teacher~\citeyearpar{unbiasedteacher}, Instant-Teaching~\citeyearpar{instantteaching} and Humble teacher~\citeyearpar{tang2021humble} update the teacher model online with an evolving student model in a similar way of Mean Teacher~\citeyearpar{meanteacher}. Instant-Teaching~\citeyearpar{instantteaching} and ISMT~\citeyearpar{ismt} further explore training an ensemble of two model backbones/heads like Co-teaching~\citeyearpar{coteaching} for better performance. A multi-phase learning method is also introduced in~\citeyearpar{multiphaselearning} to combat the noise in pseudo labels. 
While semi-supervised object detection performance has been steadily improved, most current approaches directly leverage recent advances on semi-supervised image classification for object detection. In contrast, we investigate and address the unique challenge of semi-supervised object detection---injecting localization precision to generate better boxes and dynamically adjusting pseudo label threshold to combat class imbalance.

\section{Conclusion}

In this paper, we rethink the use of pseudo labels for semi-supervised object detection (SSOD), and equip pseudo labels to be certainty-aware so as to address the lack of localization confidence when generating pseudo labels and the amplified class imbalance. We presented certainty-aware pseudo labeling considering both classification and localization quality by formulating box localization as a classification problem. Conditioned on the quality scores, the pseudo labels are filtered by dynamically derived thresholds and the losses are re-weighted in a class-specific manner, in pursuit of improved localization quality and balanced network learning for SSOD. Extensive experiments under multiple settings demonstrated the efficacy of our method. 

\section*{Appendix}

\subsection{Training and Testing Configuration} 

More details of training and testing on different experimental settings are provided as below: 
\begin{itemize}
    \item (1) \emph{COCO-full}: we report results under 1$\times$ and 3$\times$ training schedules, which are roughly equivalent to 12 and 36 epochs respectively. The teacher models are trained for 180k / 540k iterations, and the student models are trained using the same schedule.
    \item (2) \emph{COCO-partial}: for 1/2/5/10$\%$ settings, we train teacher models for 6k/12k/30k/60k iterations, and then train student models for 180k iterations. During testing, we report results under two score thresholds, $0.05$ and $0.001$, that are applied on final detection predictions. A lower threshold generally improves the recall through keeping more predicted boxes and thus results in slightly better performance.
    \item (3) \emph{PASCAL VOC}: we train teacher models for 10k iterations, then train student models for 90k iterations. Both single-scale training and multi-scale training results are reported. 
\end{itemize}

For ablation study and analysis, we use the $2\%$ COCO setting and a shorter 0.5$\times$ schedule due to the limited computational resources, if not mentioned otherwise.

It is worth pointing out that existing methods typically use a larger batch size than ours (16/32/64 \emph{v.s.} 8), and thus our training schedule is equal to -- sometimes shorter than -- other state-of-the-art SSOD approaches~\cite{stac,unbiasedteacher,tang2021humble,instantteaching,ismt}. 
 
\subsection{Implementation Details} 

The ResNet-50~\cite{resnet} backbone network we use is initialized from ImageNet pre-trained weights. We set $\lambda_u = 1.0$ and $\tau = 0.7$. For the localization branch, we set $K = 30$. For dynamic thresholding and re-weighting, we set $\gamma_1 = 0.05$ and $\gamma_2 = 0.6$. We train the models with 4 Nvidia 1080 Ti GPUs, using a total batch size of 8. We use SGD with an initial learning rate of $0.01$,  a weight decay of $1e-4$, a momentum of $0.9$. Learning rate is divided by 10 at the 120k/160k iteration for the 180k schedule, and likewise for other schedules. For single-scale training, the short side of image is resized to $600$ for PASCAL VOC and $800$ for COCO; for multi-scale training, the short side size is sampled from $(640, 800)$. The long side is kept no more than $1,333$ after resizing. Other details are the same as in Detectron2~\cite{wu2019detectron2}, which is used for our implementation.

\section*{Acknowledgement}

Z. Wu was supported by NSFC under Grant No. 62102092. H. Li was supported by IARPA via Department of Interior/Interior Business Center (DOI/IBC) contract number D17PC00345. 



\bibliography{reference}

\end{document}